\documentclass[onecolumn]{IEEEtran}
\usepackage[utf8]{inputenc}
\usepackage{hyperref}
\usepackage{cite}
\usepackage{amsmath,amssymb,amsfonts}
\usepackage{algorithmic}
\usepackage{graphicx}
\usepackage{textcomp}
\usepackage{xcolor}
\def\BibTeX{{\rm B\kern-.05em{\sc i\kern-.025em b}\kern-.08em
    T\kern-.1667em\lower.7ex\hbox{E}\kern-.125emX}}
\usepackage{hyperref}
\usepackage{booktabs}
\usepackage{float}
\usepackage{enumitem}
\usepackage{dblfloatfix}

\begin{document}
\title{EMMI: Edge Multi-Modal Intelligence for Communication-Efficient MLLM Inference via Fused Representation Compression
\thanks{Identify applicable funding agency here. If none, delete this.}
}
\author{
\IEEEauthorblockN{Motahare Mounesan and Irfan Khan}
\IEEEauthorblockA{\\
\textit{Texas A\&M University}\\
\{motahare, irfankhan\}@tamu.edu}
}


\maketitle
\begin{abstract}
Recent advances in multimodal large language models (MLLMs) have opened new opportunities for edge intelligence by enabling reasoning across heterogeneous sensor modalities, such as vision, text, and telemetry data. However, deploying these capabilities on resource-constrained edge platforms remains challenging due to the substantial computational, memory, and communication demands of modern MLLMs. Rather than transmitting raw sensor observations or partitioning neural networks at intermediate layers, Edge Multi-Modal Intelligence (EMMI) communicates a compact representation between edge devices and server resources, enabling communication-efficient edge MLLM inference. To achieve this, EMMI performs modality-specific encoding, cross-modal representation fusion, and learned compression at the edge, transmitting only a compact latent representation to server-side resources for high-capacity MLLM reasoning. 
This representation-centric design reduces communication overhead, preserves local data privacy, and provides a fixed-size interface between heterogeneous edge devices and server-side MLLMs. Evaluation on a representative multimodal benchmark demonstrates that EMMI can reduce the communication payload by 32× while maintaining comparable downstream accuracy, resulting in up to a 3.4× reduction in estimated end-to-end inference latency under bandwidth-constrained edge conditions.
\end{abstract}


\section{Introduction}
As edge systems become increasingly autonomous, their effectiveness depends on the ability to reason across heterogeneous modalities under strict resource constraints. In many operational environments, actionable insight requires jointly interpreting visual observations, sensor measurements, telemetry, and contextual data. This reliance on multi-modal reasoning has created growing interest in intelligent systems capable of integrating diverse information sources to support timely and context-aware decision-making at the edge. Recent advances in Multimodal Large Language Models (MLLMs) provide a promising foundation for achieving such capabilities by enabling unified processing of heterogeneous inputs.


Unlike conventional lightweight multi-modal models designed for specific tasks, MLLMs can capture higher-order relationships among heterogeneous inputs and perform contextual cross-modal reasoning within a unified architecture. These capabilities enable more flexible and generalizable decision-making across complex environments. However, their large computational, memory, and communication requirements create a significant deployment gap between capabilities of modern MLLMs and the resources available on edge platforms. 

To enable MLLM capabilities on edge platforms, existing deployment strategies have explored reducing model complexity, utilizing remote compute resources, and minimizing communication costs. However, these approaches address individual resource constraints while leaving the fundamental challenge of efficient multi-modal reasoning under edge limitations unresolved. Reducing model size can limit the expressive capability of MLLMs, while offloading computation or transmitting intermediate representations can introduce communication overhead and latency. Therefore, enabling practical MLLM-based intelligence at the edge requires a deployment strategy that preserves task-relevant multimodal information while adapting computation and communication to edge constraints.

Bridging this deployment gap requires careful coordination between computation and communication across edge and server resources. Direct offloading of multi-modal inputs can reduce local processing requirements but introduces significant communication overhead when transmitting high-volume sensor streams. On-device optimization techniques reduce resource consumption but often sacrifice the reasoning capability and generality that make MLLMs valuable. Existing split-computing approaches distribute model execution across edge and server devices; however, they may still require frequent exchanges of intermediate representations during inference, limiting their effectiveness under bandwidth and latency constraints. These limitations indicate that efficient edge deployment of MLLMs requires more than reducing model size or shifting computation, but rather a strategy that preserves cross-modal reasoning capability while transforming how multi-modal information is represented and communicated between edge devices and remote resources.

To address these challenges, we propose EMMI, an Edge Multi-Modal Intelligence architecture that decouples multi-modal representation generation from high-capacity MLLM reasoning across edge and server resources. Instead of transmitting raw sensor streams or relying on conventional split execution with frequent intermediate feature exchanges, EMMI performs modality-specific encoding and cross-modal fusion at the edge before communication, generating a unified multimodal representation that is subsequently compressed and transmitted to server resources. By performing multimodal interaction before transmission, EMMI changes the communication boundary from raw inputs, modality-specific features, or model activations to a compact unified representation for server-side reasoning. This representation-centric design reduces communication overhead while preserving task-relevant information and providing a fixed-size interface between heterogeneous edge devices and server-side reasoning models.


The remainder of this paper is organized as follows. Section II presents related work on MLLMs and edge intelligence. Section III describes the proposed EMMI framework and system architecture. Section IV presents the experimental methodology and evaluation setup. Section V discusses the results and analysis. Finally, Section VI concludes the paper.

\section{RELATED WORK}
\noindent\textbf{Multimodal Intelligence under Edge Constraints.} Traditional multimodal systems typically employ lightweight, task-specific architectures for applications such as human activity recognition~\cite{chung2019sensor,multilevel_har_fusion} and onboard aerial perception~\cite{yolov8_jetson_drone_2025}. Recent multimodal large language models (MLLMs), including Flamingo~\cite{alayrac2022flamingo}, BLIP-2~\cite{li2023blip2}, and LLaVA~\cite{liu2023llava}, instead support general-purpose reasoning across heterogeneous inputs. However, their computational and memory requirements limit deployment on resource-constrained edge platforms~\cite{yao2025efficientgpt4v,lin2025llm6gedge}. Recent compact MLLMs target this gap by reducing model size and computational requirements~\cite{yao2025efficientgpt4v,jin2025efficientmllm}, but remain constrained relative to full-capacity MLLMs~\cite{jin2025efficientmllm,qu2025mobileedge}. These limitations motivate approaches that preserve high-capacity multimodal reasoning while adapting computation and communication to edge constraints.
\noindent \textbf{On-Device MLLM Optimization.} Existing efforts to enable MLLM deployment at the edge reduce model computation and memory through three complementary directions: \textit{hardware acceleration}, using specialized architectures such as NPUs and processing-in-memory designs~\cite{chen2026p3llm}; \textit{model compression}, including quantization~\cite{lin2024awq}; and \textit{efficient MLLM architectures}, which redesign models for smaller footprints and improved deployability~\cite{chu2024mobilevlm,zhou2024tinyllava,yao2024minicpmv}. While these approaches improve edge efficiency, they remain constrained by the trade-off between model efficiency and high-capacity multimodal reasoning~\cite{jin2025efficientmllm,qu2025mobileedge}. \emph{EMMI instead preserves high-capacity MLLM reasoning while shifting the communication boundary toward compact multimodal representations generated at the edge.}


\noindent \textbf{Split Computing and Collaborative Inference.} Split computing enables edge--server collaborative inference by partitioning model execution across resource-constrained devices and remote servers~\cite{kang2017neurosurgeon,li2020edgent, effect, edgerl, inferedge, vaednn}. While early approaches focused on reducing communication overhead for deep neural network inference through efficient partitioning and activation transfer~\cite{matsubara2022split}, recent efforts have extended collaborative inference to large language models and multimodal systems. LLM inference frameworks explore adaptive partitioning, speculative execution, and collaborative serving to reduce the cost of model execution across heterogeneous resources~\cite{park2025specedge,zhang2025edgeshard,sled2025}. Multimodal systems further distribute modality processing and language inference across edge--cloud resources~\cite{yuan2025distmllm}. However, existing collaborative inference approaches primarily optimize \textit{where} computation is performed, while communication remains tied to intermediate activations, modality-specific features, or model states. These representations remain dependent on the underlying model architecture and input modalities, limiting scalability as multimodal systems incorporate increasingly diverse sensor streams. \emph{EMMI instead changes \textit{what} crosses the edge--server boundary by generating and compressing a unified multimodal representation at the edge and transmitting a compact latent representation for server-side MLLM reasoning.}

\noindent\textbf{Communication-Efficient Representation Transfer.} Beyond computation placement, edge MLLM deployment faces a communication challenge:
intermediate representations exchanged between the edge and server can still introduce substantial transmission overhead. Existing approaches reduce this overhead through representation compression and semantic-aware transmission. Feature compression methods such as BottleNet++~\cite{shao2020bottlenetpp} compress intermediate activations, while task-oriented communication frameworks learn compact representations relevant to downstream objectives~\cite{shao2022taskoriented,shao2023multidevice}. Learned compression techniques further improve transmission efficiency using neural compression models~\cite{balle2018variational}. Recent work extends communication-efficient inference to multimodal LLMs. TOFC~\cite{yuan2025tofc} reduces visual feature transmission through feature merging and entropy modeling in a device-edge co-inference framework. However, such approaches typically compress modality-specific representations and perform multimodal interaction after transmission, limiting their ability to exploit cross-modal dependencies at the edge. \emph{EMMI extends these efforts by changing the transmitted representation itself. Instead of sending raw sensor inputs or separate modality-specific features, EMMI performs cross-modal fusion at the edge and generates a unified latent representation for server-side MLLM reasoning. This design provides a fixed-size communication interface between heterogeneous edge sensors and high-capacity MLLMs while preserving the flexibility required for open-ended multimodal reasoning.}


\section{System Model and Solution Approach}
In this section, we introduce the edge--server system model and the design of EMMI for communication-efficient MLLM inference under edge constraints.

\begin{figure*}[t]
    \centering
    {
        \includegraphics[width=\textwidth]{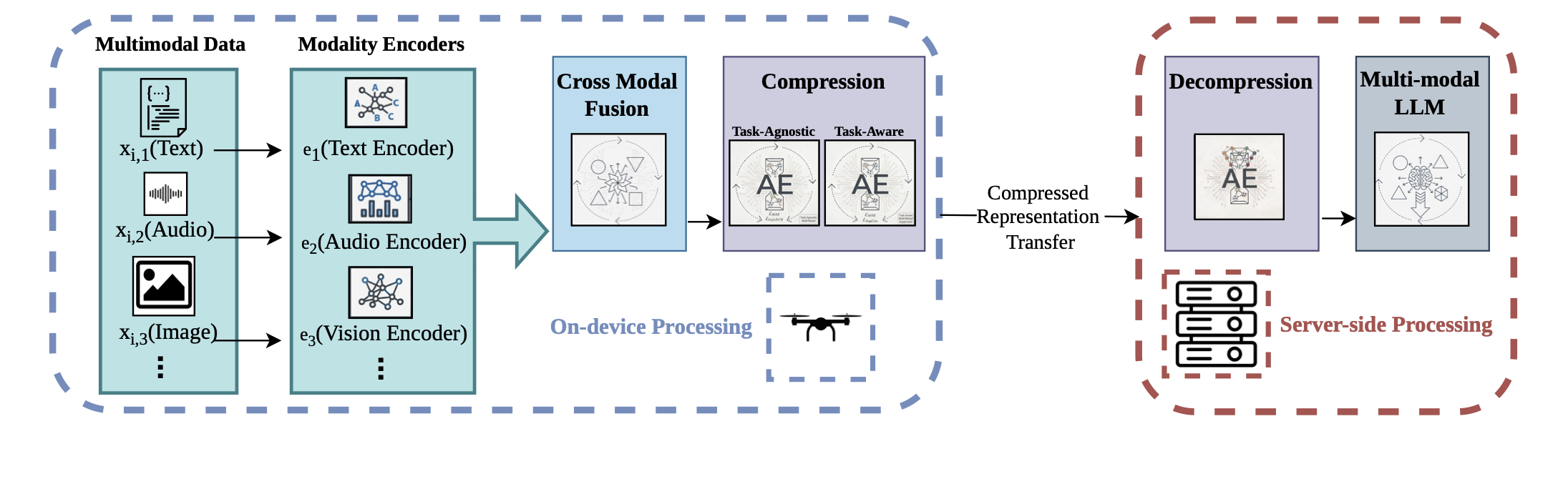}
        \vspace{-25pt}
        \caption{EMMI pipeline: encode, fuse, and compress at the edge;
        decompress, and reason at the server.}
        \label{fig:pipeline}
    }
\end{figure*}

\subsection{System Model}
\label{sec:system}
We consider a distributed edge--server multimodal intelligence system consisting
of one or more resource-constrained edge platforms connected through
bandwidth-limited wireless links to remote server infrastructure equipped with
high-capacity multimodal large language models (MLLMs). Edge platforms collect
and perform local processing on heterogeneous sensor observations, while
server-side resources provide computationally intensive multimodal reasoning. Let $\mathcal{V}=\{v_1,v_2,\dots,v_N\}$ denote a set of edge devices, where
$N$ represents the number of participating edge platforms. Each edge device
collects observations from multiple modalities at time step $t$, represented as:
\begin{equation}
X_i^t=\{x_{i,1}^t,x_{i,2}^t,\dots,x_{i,M_i}^t\},
\end{equation}

\noindent where $M_i$ denotes the number of available modalities at edge device $i$ and $x_{i,m}^t$ represents the observation from modality $m$. The system aims to obtain MLLM-based reasoning results from heterogeneous observations collected at edge devices. Since high-capacity MLLMs require substantial computational and memory resources, inference is performed collaboratively between edge platforms and remote servers. This requires transferring information from edge observations through
wireless links for server-side MLLM processing. Accordingly, end-to-end inference latency consists of edge processing ($T_{\mathrm{edge}}$), communication ($T_{\mathrm{comm}}$),
and server-side reasoning ($T_{\mathrm{server}}$ ):
\begin{equation}
T_{\mathrm{total}} =
T_{\mathrm{edge}} +
T_{\mathrm{comm}} +
T_{\mathrm{server}},
\end{equation}


The communication latency depends on the transmitted data size $S_{\mathrm{tx}}$
and available bandwidth $B$:
\begin{equation}
T_{\mathrm{comm}} =
\frac{S_{\mathrm{tx}}}{B},
\end{equation}

Therefore, the system
must balance the computational workload assigned to edge and server resources
while minimizing communication overhead under edge resource constraints.


\subsection{EMMI Framework}
\label{sec:emmi}

To enable efficient MLLM inference under the system model above, EMMI adopts a representation-centric edge--server architecture in which edge devices generate compact multimodal representations for server-side reasoning, rather than transmitting raw sensor streams or architecture-dependent activations. This design separates resource-constrained edge processing from compute-intensive MLLM reasoning while reducing communication overhead and exposure of raw sensor data. Figure~\ref{fig:pipeline} illustrates the overall architecture.

\noindent \textbf{I) Edge-side Multimodal Representation Generation. }The edge pipeline consists of three components: encoding, cross-modal fusion, and representation compression. Together, they generate the multimodal representation that serves as the communication interface between edge and server resources.

    \noindent\textbullet\textit{ Encoding:} Each modality observation is processed independently through a
    modality-specific encoder $e_m(\cdot)$, transforming  raw observations
    into intermediate feature representations:
    \begin{equation}
    h_{i,m}^{t}=e_m(x_{i,m}^{t}),
    \end{equation}
    
    \noindent where $x_{i,m}^{t}$ denotes observation from modality $m$ at edge device
    $i$ at time $t$, and $h_{i,m}^{t}$ denotes the resulting representation.

    \noindent\textbullet\textit{ Cross-modal Fusion: }The modality feature representations are combined through a cross-modal
    fusion operation to construct a unified multimodal representation:
    \begin{equation}
    f_i^t=A(h_{i,1}^{t},h_{i,2}^{t},\dots,h_{i,M_i}^{t}),
    \label{eq:alignment}
    \end{equation}
    
    \noindent where $A(\cdot)$ denotes cross-modal fusion operation and $f_i^t$
    represents resulting multimodal feature representation. By performing
    multimodal interaction before communication, EMMI captures relationships among
    heterogeneous inputs rather than transmitting independent modality-specific
    features. This enables the transmitted representation to retain cross-modal
    information relevant to downstream MLLM reasoning.
 
    \noindent\textbullet\textit{ Representation Compression: }The fused multimodal representation ($f_i^t$) is transformed into a compact
    representation:
    \begin{equation}
    z_i^t=g_{\phi}(f_i^t),
    \end{equation}
    
    \noindent where $g_{\phi}(\cdot)$ represents edge-side compression module, and $z_i^t$
    denotes the transmitted representation. The compression reduces representation dimensionality while preserving information relevant to downstream MLLM reasoning, thereby reducing communication overhead and limiting exposure of raw sensor observations.

\noindent \textbf{II) Server-side MLLM Reasoning. }The server pipeline consists of two components: representation decompression and MLLM reasoning. They transform the transmitted compact representation into the final task-specific inference result.

    \noindent\textbullet\textit{ Representation Decompression: }The server reconstructs the multimodal representation ($z_i^t$) from the received compact representation using a server-side decompression module:
    \begin{equation}
    \hat{f}_i^t=d_{\psi}(z_i^t),
    \end{equation}
    
    \noindent where $d_{\psi}(\cdot)$ denotes the server-side decompression module and $\hat{f}_i^t$ denotes the recovered multimodal representation.
    
    \noindent\textbullet\textit{ MLLM Reasoning: }The recovered multimodal representation ($\hat{f}_i^t$) is projected into the MLLM embedding space through a server-side projection module:
    \begin{equation}
    u_i^t=p_{\theta}(\hat{f}_i^t),
    \end{equation}
    
    \noindent where $p_{\theta}(\cdot)$ denotes the server-side projection module and $u_i^t$ denotes the resulting MLLM-compatible representation. The MLLM processes $u_i^t$ for multimodal reasoning, followed by a task-specific prediction head to generate the final inference result. The projection module and prediction head are configured for the downstream task, while edge-side representation generation remains unchanged. Thus, MLLM inference can be performed without access to the original sensor observations or modality-specific intermediate features.
    

This representation-centric design provides two system-level benefits. \textit{First}, EMMI keeps raw sensor observations local to the edge by transmitting only compact multimodal representations instead of raw sensor data over the edge--server link, reducing the exposure of raw sensor observations while preserving information required for server-side reasoning. \textit{Second}, although the compressed representation could be directly connected to the MLLM reasoning module, EMMI decompresses the received representation into the original fused representation space before MLLM inference. This decouples the compression mechanism from the downstream reasoning module, allowing different compression techniques or compression ratios to be adopted without modifying the reasoning module itself. Thus, the compression scheme can be changed without requiring retraining of the server-side reasoning module. The reconstructed representations can also be efficiently batched for server-side MLLM processing, enabling parallel inference and amortizing the computational cost of MLLM reasoning across requests.


\subsection{Proposed Method}
\label{sec:method}

This section describes the key design choices that enable communication-efficient multimodal inference in EMMI, focusing on cross-modal fusion before transmission and compression of the resulting representation under edge communication constraints.

\noindent \textbf{Cross-Modal Fusion.} The representation exchanged between edge and server determines the information available for downstream MLLM reasoning. EMMI performs cross-modal fusion before compression to construct a unified representation that captures both modality-specific characteristics and cross-modal interactions, reducing the need to transmit separate modality representations to the server.

EMMI instantiates the fusion function $A(\cdot)$ using pairwise interaction-based fusion, which combines modality-specific representations with explicit cross-modal interaction features. For two modalities, the fused representation is formulated as:
\begin{equation}
f_i^t=
\left[
h_{i,m}^t;
h_{i,n}^t;
\left|h_{i,m}^t-h_{i,n}^t\right|;
h_{i,m}^t\odot h_{i,n}^t
\right],
\end{equation}
\noindent where $|\cdot|$ denotes element-wise absolute difference and $\odot$ denotes element-wise multiplication. The concatenated modality features retain individual modality information, while the difference and multiplicative terms capture complementary and correlated characteristics between modalities. For more than two modalities, the same pairwise interaction operation is applied to each modality pair, and the resulting pairwise features are aggregated to form the unified multimodal representation.



\noindent \textbf{Representation Compression. } EMMI considers two compression training strategies based on the availability of downstream task supervision. When task labels are unavailable or a reusable compressed representation is desired, a task-agnostic model learns a general-purpose multimodal representation without relying on a specific inference objective. This decouples the edge-side compression module from downstream applications, allowing the same compressed representation to be reused across tasks without task-specific retraining. When task labels are available, a task-aware model incorporates supervision to optimize the compressed representation for the target inference objective while maintaining the same lightweight deployment pipeline. Both strategies operate on the same fused multimodal representation and share the same edge-side compression architecture, differing only in their training objectives.

For \textit{task-agnostic compression}, EMMI jointly optimizes reconstruction fidelity and cross-modal alignment:

\begin{equation}
\mathcal{L}_{\text{TA}}
=
\underbrace{\mathcal{L}_{\text{MSE}}(d_{\psi}(z),f)}_{\text{reconstruction}}
+
\lambda
\underbrace{\mathcal{L}_{\text{InfoNCE}}(z,\{h_m\}_{m=1}^{M})}_{\text{cross-modal alignment}},
\label{eq:ta}
\end{equation}

\noindent where $\mathcal{L}_{\text{MSE}}$ denotes the mean squared error reconstruction loss and $\mathcal{L}_{\text{InfoNCE}}$ denotes the contrastive alignment objective \cite{oord2018representation}. The reconstruction term minimizes distortion between the fused representation and its reconstruction, encouraging the latent space to retain multimodal information. The cross-modal alignment term encourages the compressed representation to remain aligned with the individual modality embeddings without downstream task labels. Together, these objectives learn a task-independent latent representation that preserves both multimodal information and cross-modal structure.

When downstream task labels are available, EMMI uses \textit{task-aware compression} to optimize the compressed representation for the target inference objective by replacing the task-independent alignment objective with supervised contrastive learning:

\begin{equation}
\mathcal{L}_{\text{TC}}
=
\underbrace{\mathcal{L}_{\text{MSE}}(d_{\psi}(z),f)}_{\text{reconstruction}}
+
\lambda
\underbrace{\mathcal{L}_{\text{SupCon}}({z_i},{y_i})}_{\text{task-specific alignment}},
\label{eq:tc}
\end{equation}


\noindent where $y_i$ denotes the downstream task label. The reconstruction term preserves information from the fused representation, while the supervised contrastive objective shapes the compressed latent space according to the target task by encouraging samples with similar labels to become closer and samples with different labels to separate. Task supervision is introduced only through the contrastive objective during training, preserving the lightweight edge-side compression and transmission pipeline while producing a representation optimized for the downstream inference objective.

\vspace{-4pt}
\section{Evaluation}
\label{sec:eval}
\vspace{-2pt}
We evaluate EMMI by first examining how effectively its compression strategies preserve multimodal information, and then assessing the resulting communication and inference efficiency. Specifically, we compare the task performance of compressed representations across methods and measure communication overhead and end-to-end latency under constrained network conditions.

\vspace{-7pt}
\subsection{Experimental Setup}
\noindent\textbf{Dataset and Multimodal Encoders.}
We use MS-COCO image-caption pairs~\cite{chen2015microsoft} as a representative vision-language benchmark to evaluate EMMI's multimodal representation and compression capabilities. The dataset is used for image-text matching. We evaluate two vision-language encoders representing different efficiency regimes: CLIP ViT-B/32 and its distilled lightweight variant MobileCLIP2-S0. CLIP ViT-B/32 uses an 87M-parameter image encoder, while MobileCLIP2-S0 reduces the encoder size to 11.4M parameters. Both encoders produce 512-dimensional embeddings, enabling comparison between full-size and lightweight vision-language models under communication-efficient compression.

\noindent\textbf{Compression Configuration.}
All evaluated compression approaches target a 64-dimensional latent representation.

\noindent\textbf{Server-side Inference.}
EMMI uses a frozen LLaVA-1.5-7B language-model backbone (6.6B parameters) as the server-side language-model backbone. A lightweight trainable two-layer MLP maps the received edge representation to eight soft-prompt tokens in LLaVA's embedding space. The frozen LLaVA backbone processes these tokens, and its final hidden representation is passed to a binary classification head.


\noindent\textbf{Baselines.}
We compare the proposed compression strategies against an uncompressed representation transfer baseline and standard dimensionality reduction and representation compression methods. The uncompressed baseline directly transmits the original fused multimodal representation and provides a reference for performance without compression. Additional baselines include principal component analysis (PCA)~\cite{hotelling1933analysis}, partitioned PCA (BlockPCA) applying PCA independently to separate blocks of the fused representation, linear discriminant analysis (LDA)~\cite{fisher1936use}, and variational autoencoder (VAE)~\cite{kingma2013auto}.

\noindent\textbf{Training Protocol.}
Training follows a two-stage decoupled protocol. In the first stage, the edge-side compression module is trained independently of the server-side inference module using either task-agnostic or task-aware supervision, depending on the compression strategy. The trained compression module is then used to generate latent representations for the training, validation, and test sets. In the second stage, the server-side projection head and classification head are trained using the cached latent representations, while the edge-side compression module remains frozen. The pretrained vision-language encoders and LLaVA backbone remain frozen throughout all experiments. We use a 72K/5K/5K train/validation/test split, with model selection based on validation loss.

\noindent\textbf{Evaluation Metrics.}
We evaluate EMMI using four metrics. \textit{Task accuracy} is the percentage of correctly classified image-text pairs on the test set. \textit{Compression ratio} is the ratio between the dimensionality of the original representation and the compressed representation. \textit{Communication payload} is the number of bytes required to transmit the compressed representation between edge and server. \textit{End-to-end latency} is the total inference time including edge-side representation generation, compression, transmission, and server-side inference.

\noindent\textbf{Evaluation Environment.}
We evaluate EMMI from both machine learning and system perspectives. Accuracy and compression effectiveness experiments, as well as server-side inference latency, are conducted on an NVIDIA Quadro RTX 6000 (24GB) to enable controlled comparison across models and compression approaches. Edge-side computational overhead is evaluated separately using a single CPU thread on an $x86\_64$ compute node, providing a reproducible proxy for resource-constrained execution.

\vspace{-5pt}
\subsection{Results and Discussion}
\label{sec:results}
We next present the experimental results and discuss the implications of EMMI's compression strategies for multimodal inference and edge--server efficiency.

\noindent\textbf{Compression Performance. }We first evaluate whether EMMI preserves task-relevant information under aggressive representation compression. All compression methods reduce the fused multimodal representation to a 64-dimensional latent representation, reducing the communication payload from 8,192B to 256B (32$\times$ reduction). We compare conventional representation compression methods against the two proposed EMMI strategies: the \emph{Task-Agnostic Multimodal Autoencoder (TaskAgnosticAE))}, which learns a reusable latent representation without downstream task labels, and the \emph{Task-Aware Supervised Contrastive Autoencoder (TaskAwareAE)}, which incorporates task supervision during training to optimize the compressed representation for the downstream inference task.

\begin{table}[t]
    \centering
    \begin{tabular}{lccc}
        \toprule
        \textbf{Method} & \textbf{Supervision} & \textbf{CLIP} & \textbf{MobileCLIP} \\
        \midrule
        No Compression & --- & 97.92 & 98.41 \\
        \midrule
        \multicolumn{4}{c}{\textit{Conventional Compression}} \\
        PCA      & -   & 96.88 & 57.35 \\
        AE       & -   & 93.40 & 52.01 \\
        VAE      & -   & 96.75 & 69.87 \\
        BlockPCA & -   & 97.12 & 97.85 \\
        LDA      & Labels & 97.78 & 98.33 \\
        \midrule
        \multicolumn{4}{c}{\textit{Proposed EMMI Compression}} \\
        TaskAgnosticAE & Pairs & 94.09 & 90.30 \\
        \textbf{TaskAwareAE} & Labels & \textbf{98.08} & \textbf{98.32} \\
        \bottomrule
    \end{tabular}
    \vspace{5pt}
    \caption{Task performance after 32$\times$ compression ($d=64$).}
    \label{tab:compression}
    \vspace{-25pt}
\end{table}

\begin{figure*}[!t]
    \centering
    \includegraphics[width=\textwidth]{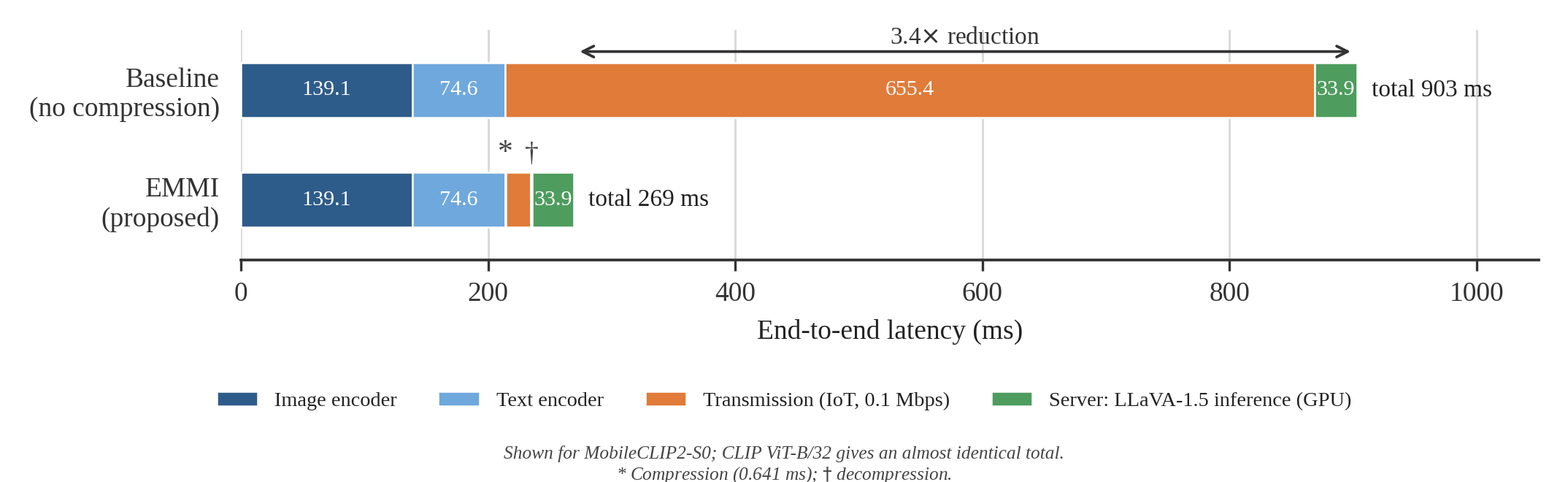}
    \vspace{-8pt}
    \caption{Estimated end-to-end latency breakdown for MobileCLIP2-S0 under a bandwidth-constrained IoT scenario (0.1~Mbps).}
    \label{fig:end_to_end}
\end{figure*}

Table~\ref{tab:compression} shows that compression performance varies substantially across representations and compression objectives. For CLIP, several conventional methods retain high task performance, whereas MobileCLIP is much more sensitive to the choice of compression method: PCA, AE, and VAE reduce accuracy by 41.06, 46.4, and 28.5 percentage points, respectively, while BlockPCA and LDA remain near the uncompressed baseline. This variation indicates that the effectiveness of a fixed compression level depends strongly on the underlying representation and compression objective, highlighting the challenge of preserving task-relevant information across different multimodal representations.



The EMMI results show that learned compression can preserve task-relevant information across different multimodal representations. The TaskAgnosticAE achieves 94.09\% accuracy for CLIP and 90.30\% for MobileCLIP without using task labels during training, enabling a reusable compressed representation across downstream applications. However, BlockPCA---a closed-form method that also requires no task labels and no training---achieves 97.12\% for CLIP and 97.85\% for MobileCLIP, indicating that unsupervised, structure-aware compression can outperform TaskAgnosticAE on this task. The TaskAwareAE incorporates task supervision through supervised contrastive learning and achieves 98.08\% accuracy for CLIP and 98.32\% for MobileCLIP, matching the uncompressed baselines within 0.16 percentage points while reducing the communication payload by 32$\times$. Notably, TaskAwareAE and LDA---a closed-form supervised baseline achieving 97.78\%/98.33\%---both substantially outperform generic compression methods (PCA, AE, and VAE) on MobileCLIP, where these methods suffer substantial degradation.

Overall, these results indicate that aggressive representation compression is not determined solely by dimensionality reduction. The compression objective should account for the structure and task relevance of the multimodal representation, particularly for compact encoders where generic compression can discard information critical to downstream inference.

\noindent\textbf{Representation and Communication Efficiency. }We next evaluate the system-level impact of EMMI's representation-centric design, focusing on representation-processing overhead, communication cost, and end-to-end latency. All edge-side latency measurements are wall-clock measurements over 200 runs with 30 warm-up iterations using a single CPU thread on an $x86\_64$ compute node, providing a reproducible proxy for resource-constrained execution.

Figure~\ref{fig:end_to_end} illustrates the resulting system-level effect under a representative bandwidth-constrained IoT scenario (0.1Mbps). Without compression, network transmission is the dominant component of end-to-end latency. EMMI reduces the transmitted representation from 8,192B to 256B, reducing transmission latency from 655.4ms to 20.5ms while adding only $0.641$ms of compression overhead. Consequently, under this IoT bandwidth condition, the dominant bottleneck shifts away from communication toward computation, reducing the estimated end-to-end latency from approximately 903ms to 269ms, a 3.4$\times$ improvement.

\begin{table}[h]
    \centering
    \begin{tabular}{lc}
        \toprule
        Component & Latency (ms) \\
        \midrule
        Cross-modal fusion (2048-dim) & $0.013 \pm 0.000$ \\
        \hline
        PCA & $0.011 \pm 0.001$ \\
        AE & $0.577 \pm 0.004$\\
        VAE & $0.187 \pm 0.003$\\
        BlockPCA & $0.014 \pm 0.001$ \\
        LDA & $0.035 \pm 0.001$\\
        TaskAgnosticAE / TaskAwareAE & $0.641 \pm 0.005$ \\
        \bottomrule
    \end{tabular}
    \vspace{5pt}
    \caption{Representation-processing latency (ms), mean $\pm$ std.}
    
    \label{tab:compression_latency}
    
\end{table}

Table~\ref{tab:compression_latency} shows that representation processing adds negligible edge-side overhead. Cross-modal fusion requires only $0.013$ms, while all evaluated compression methods require less than $1$ms, with AE-based compression requiring the most, at $0.641$ms. Thus, EMMI does not introduce a new computational bottleneck at the edge. For comparison, modality encoding accounts for over 99\% of edge-side computation, making the sub-millisecond representation-processing overhead negligible in the overall edge pipeline.

The extent of this benefit depends on network bandwidth. Table~\ref{tab:latency_tx} quantifies transmission latency for different representation sizes across representative network conditions. At 0.1Mbps, the 32$\times$ payload reduction saves approximately 635ms per inference. The corresponding savings decrease to 6.4ms at 10Mbps LTE, 1.26ms at 50Mbps WiFi, and 0.68ms at 100Mbps 5G.

\begin{table}[h]
    \centering
        \begin{tabular}{lcccc}
        \toprule
        Payload & IoT & LTE & WiFi & 5G \\
        & 0.1Mbps & 10Mbps & 50Mbps & 100Mbps \\
        \midrule
        Uncompressed (8192B) &
        655.4 & 6.6 & 1.3 & 0.7 \\
        \textbf{Compressed, 64-dim (256B)} &
        \textbf{20.5} & \textbf{0.2} & \textbf{0.04} & \textbf{0.02} \\
        \bottomrule
        \end{tabular}
        \caption{Transmission latency across representative link bandwidths.}
        \label{tab:latency_tx}  
\end{table}

These results reveal a clear operating regime for representation-centric edge--server inference. When communication is the bottleneck, EMMI converts a large communication cost into a negligible compression cost, substantially reducing end-to-end latency without degrading task performance. As bandwidth increases, transmission becomes less significant and the latency advantage correspondingly diminishes; compression then primarily provides communication savings and a compact interface between edge and server. This demonstrates that EMMI's benefit is not simply smaller representations, but the ability to shift the system away from communication-bound inference while preserving the information required for downstream reasoning.
\section{Conclusion}
\label{sec:conclusion}
This paper presented EMMI, an edge--server framework for MLLM inference under resource and communication constraints. EMMI performs modality-specific encoding, cross-modal fusion, and learned compression at the edge, transmitting compact representations for server-side reasoning. By shifting the communication boundary from raw inputs and intermediate features to compressed representations, EMMI reduces communication overhead while preserving task-relevant information. The task-aware compression strategy uses downstream task supervision to preserve task-relevant information, maintaining performance within 0.16 percentage points of the uncompressed baselines while reducing the communication payload by 32$\times$. Under bandwidth-constrained IoT conditions, this reduction yields a 3.4$\times$ improvement in estimated end-to-end latency. These results demonstrate the importance of jointly designing multimodal representations and compression for efficient edge MLLM inference. Future work will evaluate EMMI on edge hardware, extend it to more complex modalities and heterogeneous devices, and address scenarios with non-aligned modality encoders.

\bibliographystyle{IEEEtran}
\bibliography{references}

\end{document}